\documentclass[letterpaper, 10 pt, conference]{ieeeconf}  

\IEEEoverridecommandlockouts                              

\title{\LARGE \bf Cooperative distributed model predictive control for embedded systems: Experiments with hovercraft formations
}

\usepackage{graphicx}
\usepackage[hidelinks]{hyperref}
\usepackage{amsmath,amssymb,amsfonts}
\usepackage{algorithm}
\usepackage{algpseudocode}
\usepackage[style=ieee,backend=bibtex,url=false,doi=false,isbn=false,date=year,citestyle=numeric-comp,maxbibnames=10,minbibnames=10,maxcitenames=10,mincitenames=10]{biblatex}
\bibliography{paper.bib}
\bibliography{paper2.bib}
\DeclareMathOperator*{\argmin}{argmin} 

\newtheorem{rem}{Remark}

\author{Gösta Stomberg$^{*}$, Roland Schwan$^{*}$, Andrea Grillo, Colin N. Jones and Timm Faulwasser
\thanks{$^*$ equally contributing first authors of this work.}%
\thanks{This research was supported by the Swiss National Science Foundation under the NCCR Automation (grant agreement 51NF40\_225155). This work was partially supported by the Deutsche Forschungsgemeinschaft (DFG, German Research Foundation) - project number 527447339.}%
\thanks{GS and TF are with the Institute of Control Systems, Hamburg University of Technology, 21079 Hamburg, Germany (e-mail: goesta.stomberg@tu-dortmund.de, timm.faulwasser@ieee.org). GS was with the Institute of Energy Systems, Energy Efficiency and Energy Economics, TU Dortmund University, 44227 Dortmund, Germany.}
\thanks{RS, AG, and CJ are with the Automatic Control Lab, EPFL, CH-1015 Lausanne, Switzerland (e-mail: \{roland.schwan, andrea.grillo, colin.jones\}@epfl.ch). RS is also with the Risk Analytics and Optimization Chair, EPFL, CH-1015 Lausanne, Switzerland.}
}

\begin{document}

\maketitle
\thispagestyle{empty}
\pagestyle{empty}

\begin{abstract}
This paper presents experiments for embedded cooperative distributed model predictive control applied to a team of hovercraft floating on an air hockey table.
The hovercraft collectively solve a centralized optimal control problem in each sampling step via a stabilizing decentralized real-time iteration scheme using the alternating direction method of multipliers.
The efficient implementation does not require a central coordinator, executes onboard the hovercraft, and facilitates sampling intervals in the millisecond range.
The formation control experiments showcase the flexibility of the approach on scenarios with point-to-point transitions, trajectory tracking, collision avoidance, and moving obstacles.

\end{abstract}

\section{Introduction}

Model Predictive Control~(MPC) is promising for robotics, because it explicitly accounts for actuator and safety constraints, interlaces motion planning with feedback control, and is applicable to output regulation, trajectory tracking, and path following~\cite{Katayama2023}. 
While early implementations of MPC were limited to slow dynamics because of the high computational demand, progress in real-time optimization schemes and hardware has enabled the application to fast systems such as aerial swarms and racing drones~\cite{Soria2021,Romero2022}.

Distributed optimization and Distributed MPC~(DMPC) target cyber-physical systems such as energy networks~\cite{Molzahn2017} or multi-robot systems~\cite{Shorinwa2024}, where distributed computation addresses large-scale systems, privacy, and fault-tolerance.
Distinguishing properties among DMPC variants include the cooperation among subsystems, shared cost and constraints, physical or virtual coupling, the communication architecture, and the number of communications per control step~\cite{Scattolini2009,Muller2017}.

This paper focuses on cooperative DMPC schemes where subsystems jointly solve a centralized Optimal Control Problem~(OCP) in each control step.
The framework aims to recover the strong performance of centralized MPC, but without the need for a central coordinator.
Instead, only neighbor-to-neighbor communication and local computation on the robots is required.
The theory and numerical implementation of such DMPC schemes have been extensively studied~\cite{Conte2016,Summers2012,Giselsson2014}.
However, the validation of cooperative DMPC in experiments with real hardware and distributed computation, i.e., one computer per subsystem, is scarce.

\begin{figure}\centering
	\includegraphics[width=0.8\columnwidth]{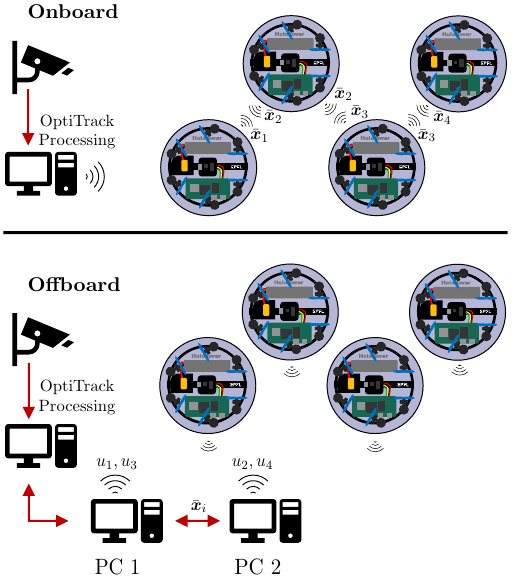}
	\caption{Schematic overview of our experimental setup. We compare onboard and offboard execution of our DMPC algorithm where coupled hovercraft exchange predicted state trajectories in each control step. For onboard execution, the position measurements are sent to the hovercraft that run the state estimation and DMPC algorithm. In offboard experiments, the observer and DMPC algorithm run on external computers that communicate via Ethernet and send the control signals to the hovercraft via Wi-Fi.}\label{fig:overview}
\end{figure}

Related work applying predictive control to multi-robot systems mostly considers approaches other than cooperative DMPC and dates back at least to~\cite{Richards2005}.
Therein, each robot exchanges predicted trajectories with neighbors and solves an individual OCP once per control step.
Similar approaches were tested with distributed computation in ~\cite{KuwataEtAl07,KanjanawanishkulZell08,Kanjanawanishkul2008b,Erunsal2024} and with centralized computation, i.e., one computer for all subsystems, in~\cite{Mehrez2017,Luis2019,Luis2020}.
To cope with limited communication, the approaches listed above modify centralized MPC for deployment on multiple robots, employ elaborate individual OCP designs treating neighboring trajectories as parameters and thus yield suboptimal performance.

In contrast, cooperative DMPC directly uses a centralized OCP, distributes the computation, and hence has the potential to provide optimal performance.
Each robot optimizes over their own and neighboring trajectories and exchanges decision variables with neighbors multiple times per control step. 
Such schemes have been tested on a four tank system~\cite{Hentzelt2014} and on mobile robots with distributed~\cite{vanParys2017,Stomberg2023} and centralized computation~\cite{Burk2021b,RosenfelderEbelEberhard22}.
However, prior experiments with distributed computation were limited to sampling frequencies below $5\,$Hz and only allowed for slow maneuvers.

This paper presents experimental results for the formation control of multiple agile hovercraft on an air hockey table, cf. Figure~\ref{fig:overview}. 
A detailed description of the hovercraft hardware can be found in~\cite{Schwan2024}.
The installed single board computer allows to run the DMPC algorithm either onboard or offboard in order to study various computational and communication scenarios such that the paper presents two main contributions.
First and in comparison to~\cite{vanParys2017}, we run multiple optimizer iterations at a faster sampling frequency of $20\,$Hz thanks to an efficient decentralized Real-Time Iteration (dRTI) scheme~\cite{Stomberg2024}.\footnote{Throughout the paper we refer to methods without central coordination as distributed, if they are control methods and as decentralized, if they are optimization methods~\cite{Scattolini2009,Nedic2018a}.} 
Second and in comparison to our prior work~\cite{Stomberg2023}, we consider more challenging maneuvers, dynamic obstacles, onboard computation, and an asynchronous implementation of the Alternating Direction Method of Multipliers~(ADMM).

The paper is structured as follows.
Section~\ref{sec:problem} introduces the formation control problem.
Section~\ref{sec:dmpc} presents the dRTI scheme for cooperative DMPC.
Section~\ref{sec:setup} describes the experimental setup and Section~\ref{sec:results} analyses the results.

\textit{Notation:}
Given a matrix $A$ and integers $i,j$, $[A]_{ij}$ is the component of $A$ at position $(i,j)$. Likewise, $[a]_i$ denotes the $i$th component of a vector $a$.
The concatenation of vectors $a$ and $b$ into a column vector is $(a,b)$. 
Given an index set $\mathcal{S}$, $(a_j)_{j \in \mathcal{S}}$ is the vertical concatenation of the vectors $a_j$ for all $j \in \mathcal{S}$.
The matrix $A = [A_{ij}]$ is the block matrix with entries $A_{ij}$ at block position $(i,j)$ for all $i,j \in \mathcal{S}$ and $B = \mathrm{diag}(R_i)$ is the block diagonal matrix with entries $R_i$ at block position $(i,i)$ for all $i \in \mathcal{S}$.  
Given a vector $a$, $A = \mathrm{diag}(a)$ is the diagonal matrix where $[A]_{ii} = [a]_i$.
The set $\mathbb{I}_{[0,N]}$ denotes the integers in range $[0,N]$ and $\mathbb{N}$ is the set of natural numbers.

\section{Problem Statement}\label{sec:problem}

We consider the formation control of a robotic swarm $\mathcal{S} = \{1,\dots,S\}$ described by an undirected time-invariant coupling graph $\mathcal{G} = (\mathcal{S},\mathcal{E})$.
Neighboring robots are connected by edges $\mathcal{E} \hspace*{-1mm}\subseteq \hspace*{-1mm}\mathcal{S} \times \mathcal{S}$ in the graph and each robot can communicate bi-directionally with its neighbors $\mathcal{N}_i \doteq \{j \in \mathcal{S} | (i,j) \in \mathcal{E}\}$.

\subsection{System Model and Formation Control}

The holonomic kinematics of each robot $i \in \mathcal{S}$ are modelled in the 2D plane with the state $x_i \in \mathbb{R}^{n_{x,i}}$, input $u_i \in \mathbb{R}^{n_{u,i}}$, and output $y_i \in \mathbb{R}^{n_{y,i}}$ with $n_{x,i} = 6$, $n_{u,i} = n_{y,i} = 3$, and
\begin{align*}
x_i &\doteq (p_{\mathrm{x},i}, p_{\mathrm{y},i}, \varphi_i, v_{\mathrm{x},i}, v_{\mathrm{y},i}, \omega_i),\\
u_i &\doteq (u_{\mathrm{x},i}, u_{\mathrm{y}_i}, u_{\varphi,i}), \quad \text{and} \quad y_i \doteq (p_{\mathrm{x},i}, p_{\mathrm{y},i}, \varphi_i).
\end{align*}
In a fixed global reference frame, $p_i = (p_{\mathrm{x},i}, p_{\mathrm{y},i})$ denotes the position of the hovercraft center, $\varphi$ is the yaw angle, $v_{\mathrm{x},i}$ and $v_{\mathrm{y},i}$ are the center velocities, $\omega_i$ is the angular velocity, $u_{\mathrm{x},i}$ and $u_{\mathrm{y},i}$ are the center acceleration, and $u_{\varphi,i}$ is the angular acceleration.
Thus, given an initial state $x_{i,0} \in \mathbb{R}^{n_{x,i}}$, we model each robot as a time-invariant discrete-time double integrator
\begin{equation*}
x_i (t+1) = A_i x_i(t) + B_i u_i(t), \quad x_i(0) = x_{i,0},
\end{equation*} where the matrices $A_i \in \mathbb{R}^{n_{x,i} \times x_{x,i}}$ and $B_i \in \mathbb{R}^{n_{x,i} \times n_{u,i}}$ are obtained via a zero-order hold discretization for the control sampling interval $\Delta t > 0$.
We note that the actual robot dynamics are more complex due to actuator nonlinearities~\cite{Schwan2024}.

The considered position-based formation control task is a setpoint stabilization problem for the centralized state $x \doteq (x_1,\dots,x_S) \in \mathbb{R}^{n_x}$, i.e., $\lim_{t \rightarrow \infty} \| x(t) - x_\mathrm{d} \| = 0$, where the desired state $x_\mathrm{d} \in \mathbb{R}^{n_x}$ encodes the formation by specifying the absolute position for each robot~\cite{Oh2015}.
We restrict the formal exposition to constant setpoints in order to simplify notation, even though our experimental results also include trajectory tracking scenarios. The extension to time-varying setpoints and trajectory tracking is straightforward.

\subsection{Obstacle and Collision Avoidance}

To avoid collisions between neighboring robots and with obstacles, we enforce collision avoidance constraints 
\begin{subequations}\label{eq:avoidance}
\begin{align}
\left\| (p_{\mathrm{x},i}, p_{\mathrm{y},i}) - (p_{\mathrm{x},j}, p_{\mathrm{y},j}) \right\|_2^2 &\geq {d^2_\mathrm{min}} \;\, \forall j \in \mathcal{N}_i, \label{eq:collavoidance}\\
\hspace*{-2mm}\left\| (p_{\mathrm{x},i}, p_{\mathrm{y},i}) - (p_{\mathrm{x},\mathrm{obs}}(t),p_{\mathrm{y},\mathrm{obs}}(t)) \right\|_2^2 &\geq {d^2_\mathrm{min}}\label{eq:obsavoidance}
\end{align} 
\end{subequations} with $d_\mathrm{min} > 0$ for all $i \in \mathcal{S}$.
Here, $p_{\mathrm{x},\mathrm{obs}}(t)$ and $p_{\mathrm{y},\mathrm{obs}}(t)$ is the obstacle position at sampling instant $t$. 

\section{Distributed Model Predictive Control Via Decentralized Real-Time Iterations}\label{sec:dmpc}

At sampling step $t$, the robots cooperatively solve the OCP

\begin{subequations}\label{ocp}
	\begin{align}
	\min_{\boldsymbol{x},\boldsymbol{u}} \sum_{i \in \mathcal{S}} &\left( \sum_{\tau=0}^{N-1} \ell_i \left({x}[\tau],{u}_i[\tau]\right) + V_{\text{f},i}({x}[N]) \right)\label{ocp:cost}\\
	\nonumber\text{subject} &\text{ to for all } i \in \mathcal{S}\\
	{x}_i[\tau+1] &= A_i{x}_i[\tau] + B_i {u}_i[\tau] \hspace*{0.33cm} \forall \tau \in \mathbb{I}_{[0,N-1]},\label{ocp:dyn}\\
	({x}_i[0],{u}_i[0]) &= (x_i(t),u_i(t)),\label{ocp:x0}\\
	{x}_i[\tau] &\in \mathbb{X}_i(t) \hspace*{2.2cm} \forall \tau \in \mathbb{I}_{[0,N]}\label{ocp:xi},\\
	{u}_i[\tau] &\in \mathbb{U}_i \hspace*{2.6cm} \forall \tau \in \mathbb{I}_{[0,N-1]}\label{ocp:ui},\\
	\left({x}_i[\tau],{x}_j[\tau]\right) &\in \mathbb{X}_{ij} \hspace{1.01cm} \forall j \in \mathcal{N}_i, \hspace*{0.1cm} \forall \tau \in \mathbb{I}_{[0,N]}.\label{ocp:xij}
	\end{align}
\end{subequations}
The decision variables ${\boldsymbol{x}} \doteq \left({x}[0],\dots,{x}[N]\right)$ and ${\boldsymbol{u}} \doteq ({u}[0],\dots,{u}[N-1])$ with $u \doteq (u_1,\dots,u_S) \in \mathbb{R}^{n_u}$ are the predicted trajectories over the horizon $N$.
Here and in the following, square brackets mark predicted variables and bold symbols denote trajectories.
The objective penalizes the deviation from the setpoint $x_\mathrm{d}$ via the convex stage costs $\ell_i : \mathbb{R}^{n_x} \times \mathbb{R}^{n_{u,i}} \rightarrow \mathbb{R}$, 
\begin{equation*}
\ell_i(x,u_i) \doteq \hspace*{-1mm}\sum_{j \in \mathcal{S}} \frac{1}{2} (x_i - x_{i,\mathrm{d}})^\top Q_{ij} (x_j - x_{j,\mathrm{d}}) + \frac{1}{2} \| u_i - u_{i,\mathrm{d}} \|_{R_i}^2
\end{equation*} and the convex terminal penalties $V_{\mathrm{f},i}: \mathbb{R}^{n_x} \rightarrow \mathbb{R}$,
\begin{equation*}
V_{\mathrm{f},i}(x) \doteq \sum_{j \in \mathcal{S}} \frac{1}{2} (x_i - x_{i,\mathrm{d}})^\top P_{ij} (x_j - x_{j,\mathrm{d}}).
\end{equation*} The centralized weight matrices $Q \doteq [Q_{ij}] \in \mathbb{R}^{n_{x} \times n_{x}}$, $R \doteq \mathrm{diag}(R_{i}) \in \mathbb{R}^{n_{u} \times n_u}$, and $P \doteq [P_{ij}] = \mathbb{R}^{n_{x}} \times \mathbb{R}^{n_{x}}$ are positive definite and $Q_{ij} = P_{ij} = 0$ if $j \notin \mathcal{N}_i \cup \{i\}$ for all $i \in \mathcal{S}$, i.e., only neighbors are coupled through the cost~\eqref{ocp:cost}.
The state constraint set
\begin{equation*}
\mathbb{X}_i \doteq \left\{x_i \in \mathbb{R}^{n_{x,i}} \; \left| \; \underline{p}_{i} \leq {p}_i \leq \overline{p}_{i} \right.,\; \eqref{eq:obsavoidance} \right\}
\end{equation*} with $\underline{p}_i,\overline{p}_i \in \mathbb{R}^2$ includes box and obstacle avoidance constraints for robot $i$.
It is time varying for moving obstacles.
The collision avoidance constraints between neighboring robots are included as coupled state constraint sets
\begin{equation*}
\mathbb{X}_{ij} \doteq \left\{ (x_i,x_j) \in \mathbb{R}^{n_{x,i}} \times \mathbb{R}^{n_{x,j}} \; \left| \;\eqref{eq:collavoidance} \right. \right\},	
\end{equation*} and the sets $\mathbb{U}_i$ restrict the commanded acceleration. 

\begin{rem}[Compensation of computational delay]\label{rem:delay}
	The initial condition on the input~\eqref{ocp:x0} compensates for any delay up to the control sampling interval, because we solve OCP~\eqref{ocp} for the input $u[1]$ to be applied in the next control step~\cite{Findeisen2006}.
	This is crucial as solving OCP~\eqref{ocp} with multiple communication rounds can be slow~\cite{Stomberg2023}. \hfill $\square$
\end{rem}

To obtain a DMPC scheme with only neighbor-to-neighbor communication, we solve OCP~\eqref{ocp} via the dRTI scheme from~\cite{Stomberg2024}.
We introduce state copies ${w}_{ji} = {x}_j$ for all $j \in \mathcal{N}_i$ which become additional decision variables of subsystem $i$ and replace ${x}_j$ in the coupled costs~\eqref{ocp:cost} and state constraints~\eqref{ocp:xij}.
Denote the predicted copy trajectories as ${\boldsymbol{w}}_i \doteq ({\boldsymbol{w}}_{ji})_{j \in \mathcal{N}_i}$ with $ {\boldsymbol{w}}_{ji} \doteq   ({w}_{ji}[0],\dots,{w}_{ji}[N])$.
We collect the decision variables in the vector $z_i \doteq ({\boldsymbol{x}}_i,{\boldsymbol{u}}_i,{\boldsymbol{w}}_i) \in \mathbb{R}^{n_i}$ for all $i \in \mathcal {S}$ and rewrite OCP~\eqref{ocp} as a non-convex partially separable Nonlinear Program~(NLP)
\begin{subequations}\label{nlp}
	\begin{align}
	\min_{z_1 ,\dots,z_S} &\sum_{i \in \mathcal{S}} f_i (z_i)\label{nlp:obj}\\
	\text{subject to }g_i(z_i) &=0, \quad h_i(z_i) \leq 0 \quad \forall i \in \mathcal{S},\label{nlp:eq}\\ 
	\sum_{i \in \mathcal{S}} E_i z_i &= 0. \label{nlp:coup}
	\end{align}
\end{subequations}
The functions $f_i: \mathbb{R}^{n_i} \rightarrow \mathbb{R}$, $g_i : \mathbb{R}^{n_i} \rightarrow \mathbb{R}^{n_{g,i}}$, and $h_i : \mathbb{R}^{n_i} \rightarrow \mathbb{R}^{n_{h,i}}$ are three times continuously differentiable and are composed of the components of OCP~\eqref{ocp}. That is, $f_i$ are the cost functions, $g_i$ include the initial conditions and system dynamics, and $h_i$ include the collision avoidance, obstacle avoidance, and box constraints.
The coupling constraint~\eqref{nlp:coup} with the sparse matrices $E_i \in \mathbb{R}^{n_c} \times \mathbb{R}^{n_i}$ enforces equivalence between the original and copied states.

The dRTI scheme executes computations on two levels, with Sequential Quadratic Programming~(SQP) iterations on the outer and ADMM iterations on the inner level. We index outer iterations with a superscript $\cdot^k$ and inner iterations by $\cdot^{k,l}$.
The outer level approximates NLP~\eqref{nlp} at a point $z^k = (z_1^k,\dots,z_S^k) \in \mathbb{R}^{n}$ as the convex Quadratic Program (QP)
\begin{subequations}\label{eq:QPadmm} 
	\begin{align}
	\min_{\substack{z_1 \in \mathbb{Z}_1^k, \dots, z_S \in \mathbb{Z}_S^k\\ \bar{z}\in \mathbb{E}} }  \sum_{i \in \mathcal{S}} & f_i^{\mathrm{QP},k}(z_i)\label{eq:QPadmmObj}\\
	\textrm{subject to } \quad  z_i - \bar{z}_i &= 0 \; | \; \gamma_i \quad \quad \forall i \in \mathcal{S}\label{eq:QPadmmCons}.	
	\end{align}
\end{subequations}
For all $i \in \mathcal S$, QP~\eqref{eq:QPadmm} includes the auxiliary decision variables $\bar{z}_i \doteq (\bar{\boldsymbol{x}}_i,\bar{\boldsymbol{u}}_i,\bar{\boldsymbol{w}}_i)\in \mathbb{R}^{n_i}$, the consensus Lagrange multipliers $\gamma_i \in \mathbb{R}^{n_i}$, the objective $f_i^{\mathrm{QP},k} \doteq (z_i-z_i^k)^{\top} H_i^k (z_i-z_i^k)/2 + \nabla f_i(z_i^k)^{\top}(z_i-z_i^k) $, and the
subsystem constraint set
\begin{align*}
\mathbb{Z}_i^k \doteq \left\{ z_i \in \mathbb{R}^{n_i} \left| \; \begin{aligned} g_i(z_i^k) + \nabla g_i(z_i^k)^{\top} (z_i-z_i^k)&= 0\\
h_i(z_i^k) + \nabla h_i(z_i^k)^{\top} (z_i-z_i^k) &\leq 0 \end{aligned} \right. \right\}.
\end{align*}
The auxiliary decision variables are coupled through the set
\begin{align*}
\mathbb{E} \doteq \left\{ \bar{z} \in \mathbb{R}^n \left| \; \sum_{i \in \mathcal{S}} E_i \bar{z}_i = 0 \right. \right\}.
\end{align*}
We use the positive-definite Gauss-Newton Hessian approximation $H_i = \nabla_{z_iz_i}^2 f_i(z_i)$ and we compute $H_i$ offline as $f_i$ is quadratic for all $i \in \mathcal{S}$.
An alternative would be to use the exact Hessian of the Lagrangian and to regularize if needed.

To obtain a decentralized optimization scheme, we solve QP~\eqref{eq:QPadmm} via ADMM by defining the augmented Lagrangian,
\begin{align*}
L^k_\rho(z,\bar{z},\gamma) &\doteq \sum_{i \in \mathcal{S}} L_{\rho,i}^k(z_i,\bar{z}_i,\gamma_i),
\end{align*} where $L_{\rho,i}^k \doteq  f_i^{\mathrm{QP},k}(z_i) + \gamma_i^\top (z_i-\bar{z}_i) + \frac{\rho}{2} \| z_i - \bar{z}_i \|_2^2$ with the penalty parameter $\rho > 0$.
The ADMM iterations read
\begin{subequations}\label{eq:ADMM}
	\begin{align}
	z_i^{k,l+1} &= \argmin_{z_i \in \mathbb{Z}_i^k} L_{\rho,i}^k(z_i,\bar{z}_i^{k,l},\gamma_i^{k,l}),\label{eq:admmZ}\\
	\bar{z}^{k,l+1} &= \argmin_{\bar{z} \in \mathbb{E}} L_{\rho}^k(z^{k,l+1},\bar{z},\gamma^{k,l}),\label{eq:admmAvg}\\
	\gamma_i^{k,l+1} &= \gamma_i^{k,l} + \rho (z_i^{k,l+1} - \bar{z}_i^{k,l+1}).\label{eq:admmD}
	\end{align}
\end{subequations}
Each subsystem executes~\eqref{eq:admmZ} and~\eqref{eq:admmD} in parallel.
QP~\eqref{eq:QPadmm} is a consensus problem and thus~\eqref{eq:admmAvg} averages the original and copied states~\cite{Boyd2011,Stomberg2022},
\begin{equation}\label{eq:avg}
\hspace*{-2mm}\bar{\boldsymbol{x}}_i^{k,l+1} \hspace*{-1mm}=\hspace*{-1mm} \frac{1}{| \mathcal N_i | \hspace*{-1mm}+\hspace*{-1mm}1} \hspace*{-1mm} \left( \hspace*{-1mm} \boldsymbol{x}_i^{k,l+1} \hspace*{-1mm}+\hspace*{-1mm} \frac{\gamma_{i,\boldsymbol{x}_i}^{k,l}}{\rho} \hspace*{-1mm}+\hspace*{-1mm}\sum_{j \in \mathcal N_i} \hspace*{-1mm} \left( \hspace*{-1mm} \boldsymbol{w}_{ij}^{k,l+1} \hspace*{-1mm}+\hspace*{-1mm} \frac{\gamma_{j,\boldsymbol{w}_{ij}}^{k,l}}{\rho} \hspace*{-1mm} \right) \hspace*{-1mm}  \right) \hspace*{-1mm}.
\end{equation} Here, $\gamma_{i,\boldsymbol{x}_i}\in \mathbb{R}^{(N+1)\cdot n_{x,i}}$ and $\gamma_{j,\boldsymbol{w}_{ij}} \in \mathbb{R}^{(N+1)\cdot n_{x,i}}$ are the Lagrange multipliers in $\gamma_i$ and $\gamma_j$ that correspond to the constraints $\boldsymbol{x}_i - \bar{\boldsymbol{x}}_i = 0$ and $\boldsymbol{w}_{ij} - \bar{\boldsymbol{w}}_{ij} = 0$ in~\eqref{eq:QPadmmCons}.
Algorithm~\ref{alg:dmpc} summarizes the cooperative DMPC scheme, where the decentralized averaging~\eqref{eq:avg} is executed in Lines~\ref{dmpc:zcomm}--\ref{dmpc:zbar}.

\begin{rem}[Asynchronous decentralized ADMM]
	As such, the considered dRTI scheme is a synchronous method. However, we implement an ADMM variant which can run asynchronously if necessary to address imperfect communication.
	Unlike synchronous ADMM, this requires the inclusion of $\gamma$ in~\eqref{eq:avg}, which can be derived from the Karush-Kuhn-Tucker system of~\eqref{eq:admmAvg}~\cite[Ch. 7]{Boyd2011}. 
	This adds only minor overhead compared to synchronous implementations, because it increases the size, but not the number, of messages in Line~\ref{dmpc:zcomm} of Algorithm~\ref{alg:dmpc}. \hfill $\square$	
\end{rem}

\begin{rem}[Closed-loop stability~\cite{Stomberg2024}]
	Algorithm~\ref{alg:dmpc} is guaranteed to locally stabilize the setpoint, if the sampling frequency and number of ADMM iterations $l_\mathrm{max}$ are sufficiently large, if the exact Hessian is used, and if OCP~\eqref{ocp} is stabilizing~\cite[Theorem 2]{Stomberg2024}. \hfill $\square$
\end{rem}

\begin{rem}[Linearized collision avoidance]
	Step~\ref{dmpc:buildQP} of Algorithm~\ref{alg:dmpc} linearizes the non-convex collision avoidance constraints~\eqref{eq:avoidance}, generalizing the on-demand collision avoidance and separating hyperplane strategies in~\cite{Luis2019,vanParys2017}. \hfill $\square$
\end{rem}

\begin{algorithm}[t]
	\caption{dRTI for cooperative DMPC on robot $i$} \label{alg:dmpc}
	\begin{algorithmic}[1]
		\State Initialize: $u_i(0), {x}_{i,\mathrm{d}}, z_i(0), \gamma_i(0)$, $k_\mathrm{max},l_\mathrm{max},\rho$
		\For{NMPC step $t = 0,1,\dots$}
		\State Estimate state $x_i(t)$ and apply input $u_i(t)$
		\State Warm start SQP $z_i^0 = z_i(t)$, $\gamma_i^0 = \gamma_i(t)$
		\For{SQP iteration $k = 0,\dots,k_\mathrm{max}-1$}
		\State Compute $\nabla f_i(z_i^k), g_i(z_i^k), \nabla g_i(z_i^k), h_i(z_i^k),$
		
		\hspace*{2.3mm} and $\nabla h_i(z_i^k)$ \label{dmpc:buildQP}
		\State Initialize ADMM $\bar{z}_i^{k,0} = z_i^k$ and $\gamma_i^{k,0} = \gamma_i^k$
		\For{ADMM iteration $l = 0,\dots,l_\mathrm{max}-1$}
		\State \hspace*{-3mm} Solve QP $z_i^{k,l+1} = \displaystyle\argmin_{z_i \in \mathbb{Z}_i^k}  L_{\rho,i}^{k}(z_i, \bar{z}_i^{k,l}, \gamma_i^{k,l})$\label{dmpc:qp}
		\State \hspace*{-3mm} Send ${\boldsymbol{w}}^{k,l+1}_{ji}$ and $\gamma_{i,\boldsymbol{w}_{ji}}^{k,l}$ to all ${j \in \mathcal{N}_i}$\label{dmpc:zcomm}
		\State \hspace*{-3mm} Receive ${\boldsymbol{w}}^{k,l+1}_{ij}$ and $\gamma_{j,\boldsymbol{w}_{ij}}^{k,l}$ from all ${j \in \mathcal{N}_i}$\label{dmpc:recv_z}
		\State \hspace*{-3mm} Compute average trajectory $\bar{\boldsymbol{x}}_i^{k,l+1}$ from~\eqref{eq:avg}
		\State \hspace*{-3mm} Send average $\bar{\boldsymbol{x}}^{k,l+1}_i$ to all $j \in \mathcal{N}_i$
		\State \hspace*{-3mm} Receive average $\bar{\boldsymbol{x}}^{k,l+1}_j$ from all $j \in \mathcal{N}_i$\label{dmpc:recv_zbar}
		\State \hspace*{-3mm} Set $\bar{\boldsymbol{w}}^{k,l+1}_{ji} \hspace*{-0.5mm}  = \hspace*{-0.5mm} \bar{\boldsymbol{x}}^{k,l+1}_j \, \forall \, j \in \mathcal{N}_i$, $\bar{\boldsymbol{u}}^{k,l+1}_i \hspace*{-0.5mm} = \hspace*{-0.5mm} {\boldsymbol{u}}^{k,l+1}_i$\label{dmpc:zbar}
		\State \hspace*{-3mm} Form $\bar{z}_i^{k,l+1} = (\bar{\boldsymbol{x}}^{k,l+1}_i, \bar{\boldsymbol{u}}^{k,l+1}_i, \bar{\boldsymbol{w}}^{k,l+1}_{i})$ 
		\State \hspace*{-3mm} $\gamma_{i}^{k,l+1} =\gamma_{i}^{k,l} + \rho (z_i^{k,l+1}-\bar{z}_i^{k,l+1})$
		\EndFor
		\State $z_i^{k+1} = \bar{z}_i^{k,l_\mathrm{max}}$, $\gamma_i^{k+1} = \gamma_i^{k,l_\mathrm{max}} $
		\EndFor
		\State Extract ${u}_i[1]$ from $z_i^{k_\mathrm{max}}$
		\State Set $u_i(t+ 1) = {u}_i[1]$ to compensate delay
		\State $z_i(t+ 1) = z_i^{k_\mathrm{max}}$, $\gamma_i(t+1) = \gamma_i^{k_\mathrm{max}}$
		\EndFor
	\end{algorithmic}
\end{algorithm}

\section{Experimental Setup}\label{sec:setup}

\subsection{Hovercraft}

Each hovercraft consists of racing drone hardware mounted on a $150\,$mm diameter foam disk, weighs $145\,$g, and can float on an air hockey table. 
Six propellers generate thrust in all directions within the x-y plane, allowing the system to be modeled as a point mass. 
Compared to the original system~\cite{Schwan2024}, we here use an improved version that is easier to assemble and has enhanced computing capabilities.
Instead of mounting the propellers on a 3D-printed ring, they are now directly attached to a printed circuit board, providing increased rigidity and improved power delivery. 
Moreover, the ESP32 has been replaced by a Radxa Zero 3W with a quad-core Arm Cortex A55 and clock speeds up to $1.6\,$GHz. 

\subsection{State Estimation}

An OptiTrack system measures the hovercraft position to millimeter accuracy using reflective markers mounted on the foam base.
To estimate velocities and disturbances, we implement a continuous-time Extended Kalman Filter (EKF) with discrete measurements~\cite{OptimalStateEstimation2006}.
A constant-disturbance model is employed, which is then integrated into the system dynamics of the OCP to achieve offset-free tracking.
In particular, the EKF estimates the extended state 
$\xi_i \doteq (x_i, d_i) \in \mathbb{R}^9$,
with the disturbance $d_i \doteq (d_{\mathrm{x},i}, d_{\mathrm{y},i}, d_{\varphi, i}) \in \mathbb{R}^3$. 
The model~\eqref{ocp:dyn} is replaced by the disturbed system dynamics
\begin{equation*}
x_i[\tau+1] = A_i x_i[\tau] + B_i u_i[\tau] + B_i d_i \quad \forall \tau \in \mathbb{I}_{[0,N-1]},
\end{equation*}
which does not change the number of decision variables as $d_i$ is a parameter in the OCP.
The EKF operates alongside the DMPC algorithm and can run either offboard or onboard.

\subsection{Communication}

We rely on the Robot Operating System (ROS2) for communication between the onboard and offboard computing devices~\cite{Macenski2022}. As outlined in Figure~\ref{fig:overview}, we have two experimental setups:
1) The state estimation and DMPC algorithm run onboard the hovercraft, and only position measurements are transmitted externally. The hovercraft communicate via Wi-Fi using the reliable transport protocol provided by ROS2. We found that the \textit{best effort} approach led to many messages being dropped. 2) In offboard experiments, the state estimation and DMPC algorithm run on two external Minisforum UM790 Pro computers connected via Ethernet. The nodes are distributed on the two computers such that communication in the DMPC algorithm has to go through the Ethernet layer to simulate proper delay, i.e., DMPC nodes of neighboring robots run on different machines.

\subsection{Distributed MPC Implementation and Design}

A C++ implementation of Algorithm~\ref{alg:dmpc} is available online.\footnote{\url{https://github.com/PREDICT-EPFL/holohover}}
The DMPC controller for each robot is implemented as a ROS node to facilitate communication.
Vectors and matrices are stored with the {Eigen} library, and {CasADi} libraries are code-generated for the OCPs to evaluate derivatives~\cite{Andersson2019}.
We use the sparse interface of the QP solver PIQP to solve the subsystem QPs in Line~\ref{dmpc:qp} of Algorithm~\ref{alg:dmpc}~\cite{Schwan2023}.
To promote a timely delivery of control signals despite spikes in communication delay, each subsystem waits at most $25\,$ms in each of Lines~\ref{dmpc:recv_z} and~\ref{dmpc:recv_zbar} of Algorithm~\ref{alg:dmpc} or until $75\,$\% of the control sampling interval has elapsed.
If not all expected messages have arrived by then, the respective step is completed asynchronously.

We consider a set $\mathcal{S} = \{1,2,3,4\}$ of hovercraft and choose a path graph as coupling graph, i.e., $\mathcal{N}_i = \{i-1,i+1\} \cap \mathcal{S}$ for all $i \in \mathcal{S}$.
We design an OCP with sampling interval $\Delta t = 50\,$ms and horizon $N = 20$ and a second OCP with $\Delta t = 150\,$ms and $N = 7$.
We set $Q_{11} = Q_{22} = Q_{33} = \mathrm{diag}(28,28,18,18,40,18)$, $Q_{44} = \mathrm{diag}(14,14,9,9,20,9)$, $Q_{ij} = -\mathrm{diag}(14,14,9,9,20,9)$ for all $j \in \mathcal{N}_i$ and $R_{ii} = \mathrm{diag}(0.1,0.1,0.1)$ for all $i \in \mathcal{S}$.
We note that the matrices $Q_{ij}$ are negative because of penalizations in the relative position error $(x_i - x_j) - (x_i^\mathrm{d} - x_j^\mathrm{d})$ and that the cost function~\eqref{ocp:cost} still is convex.
The matrices $P_{ij}$ are obtained by solving the algebraic Riccati equation for the pair $(Q_{ij},R_{ii})$ for all $j \in \mathcal{N}_i \cup \{i\}$ and for all $i \in \mathcal{S}$.
The commanded accelerations are limited to $(-5\,\text{m/s}^2,-5\,\text{m/s}^2,-15\,\text{rad/s}^2) \leq u_i \leq (5\,\text{m/s}^2,5\,\text{m/s}^2,15\,\text{rad/s}^2)$ and the position box constraints are chosen to keep a $30\,$mm safety margin to the table boundary. 
The specified minimum distance to neighbors and obstacles is $d_\mathrm{min} = 200\,$mm. For ADMM, $\rho = 4$ for setpoint stabilization and $\rho = 0.1$ for trajectory tracking.

\begin{rem}[Parameter tuning]
	We tuned $Q$ and $R$ in experiments for sufficient damping of the linear-quadratic regulator-controlled system obtained as a byproduct from the Riccati equation. We tuned ADMM in simulations and found a range of penalty parameters to work well.
	\hfill $\square$
\end{rem}

\begin{rem}[Soft state constraints]
	To ensure feasibility of the~OCP, we implement all state constraints as soft constraints~\cite{Luis2019}. 
	To this end, we add slack variables to all state inequality constraints and quadratic penalties on the slack variables to the objective.
	This is necessary because of inevitable disturbances in experiments as well as consensus errors due to the limited number of ADMM iterations.	 
	Consequently, violations of the state constraints may occur but are heavily penalized.\hfill $\square$
\end{rem}

\section{Experiments} \label{sec:results}

We test the DMPC scheme in experiments of varying difficulty including point-to-point transitions, trajectory tracking, and a dynamic obstacle. 
Video recordings of all experiments are available online.\footnote{\url{https://www.youtube.com/watch?v=-ojLJUFMong}}
To analyze strengths and weaknesses of the approach, we compare different OCP designs and, by switching between onboard and offboard computation, optimizer settings.
To evaluate control performance, we consider the averaged closed-loop cost
\[
J  \doteq \frac{1}{T_\mathrm{f}} \sum_{t = 0}^{t_n} \Delta t \cdot \ell\left(x(t),u(t)\right),
\] where $T_\mathrm{f} = 120\,$s is the duration and $t_n \doteq T_\mathrm{f} / \Delta t$ is the sample number per experiment.
Additionally, we compare the numbers of collisions and averaged constraint violations. 

\subsection{Point-to-Point Transitions without Obstacle}
We first test the tracking accuracy, collision avoidance, and cooperation among the hovercraft.
The top row of Figure~\ref{fig:rectangle} shows a sequence where the hovercraft first transition from right to left. 
Then, between $t = 1.00\,$s and $t = 1.50\,$s, the setpoint changes and the hovercraft assume a rectangular formation.
The center row shows a switch in position between neighbors, activating the collision avoidance constraints. 

Because of the low number of onboard real-time feasible ADMM iterations, there is a significant number of constraint violations and, as a result, collisions, cf. Table~\ref{tab:exp_results}.
Columns two and four report results for the same scenario with onboard computation, but different sampling interval and optimizer iterations. 
This highlights the so-called \textit{real-time dilemma} of MPC~\cite{Gros2020}: Choosing a small sampling interval and applying suboptimal feedback vs. choosing a large sampling interval and applying feedback based on outdated information.
Here, a larger sampling interval leads to a worse tracking performance, presumably because of disturbances and inaccuracies in the thrust model. 
However, there occur less collisions and constraint violations with $\Delta t = 150\,$ms, indicating a reduced consensus error due to a larger number of optimizer iterations per control step.
As expected, offboard computation shows better performance, because more ADMM iterations can run in real time.
In all three cases, the dRTI scheme executes in real time, with the exception of outliers in the onboard experiments at $\Delta t = 50\,$ms, cf. Figure~\ref{fig:boxplots}.
The time needed to construct the QP in Step~\ref{dmpc:buildQP} of Algorithm~\ref{alg:dmpc} always lies below $3.4\,$ms and $0.1\,$ms per control step for onboard and offboard experiments, respectively.

Notably, wireless communication appears to be \textit{the} bottleneck for onboard execution, because dRTI spends most of the time in ADMM waiting or communicating. 
This is in line with findings reported in~\cite{Burk2021}, but different compared to ADMM implementations solving NLPs on each subsystem via ipopt~\cite{vanParys2017,Wachter2006}.

\subsection{Point-to-Point Transitions with Obstacle}

\begin{figure*}
	\includegraphics[width = \textwidth]{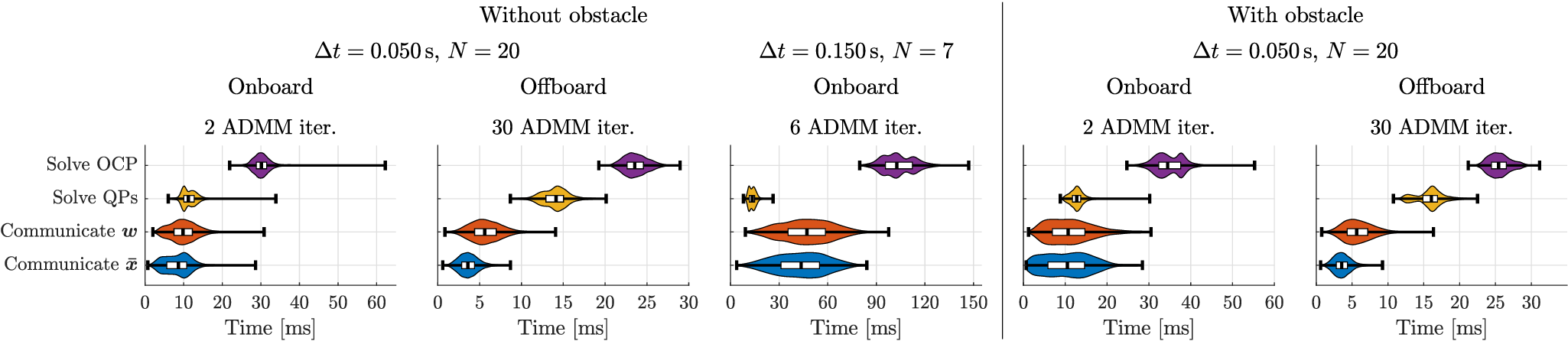}
	\caption{Optimizer execution times per MPC step for five point-to-point transition experiments. Black vertical lines indicate the minimum, median, lower and upper quartiles, and maximum and colored areas highlight probability densities. All OCP solve times lie well below the control sampling interval and can thus be compensated, except for outliers in the onboard experiments with sampling interval $\Delta t = 50\,$ms. Figure produced with daviolinplot~\cite{Karvelis2024}.}
	\label{fig:boxplots}
\end{figure*}

\begin{figure*}
	\includegraphics[width = \textwidth]{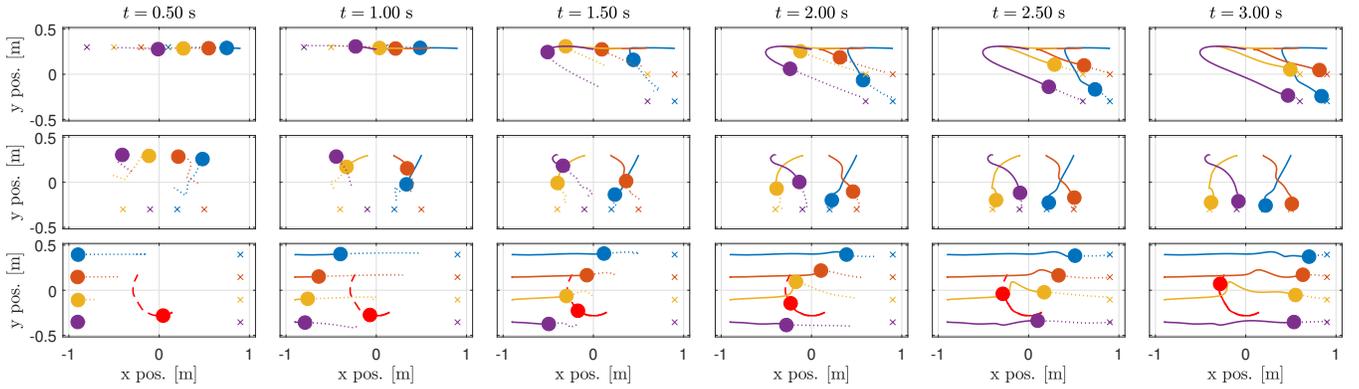}
	\caption{Hovercraft motion in the 2D plane for three maneuvers with onboard computation and sampling interval $\Delta t = 50\,$ms. Solid and dotted lines show closed-loop and predicted trajectories, respectively. Circles are the current hovercraft positions and crosses mark the setpoints. The top row depicts an abrupt setpoint change with a reconfiguration from a line into a rectangular formation. The center row shows a switch in position between neighbors while avoiding collisions. In the bottom row, the four hovercraft cross the table from left to right while dodging a dynamic obstacle shown in red. The red dashed line is the future obstacle trajectory which is unknown to the hovercraft and which is plotted for better visualization of the experiment. }
	\label{fig:rectangle}
\end{figure*}

In the second scenario, the hovercraft repeatedly cross the table from left to right and back while avoiding an obstacle.
The obstacle is static at first and then completes various circular patterns, requiring a fast sampling interval of $\Delta t = 50\,$ms.
The bottom row of Figure~\ref{fig:rectangle} depicts one sequence of this scenario with a moving obstacle and onboard computation. 
The obstacle avoidance constraints of the third hovercraft shown in yellow activate between $t = 1.00\,$s and $t = 1.50\,$s such that a collision is avoided.
However, the stark differences in the predicted trajectories among the hovercraft at $t = 0.50\,$s visualize the lack of consensus, which is the price we pay for fast control sampling.
The two plots on the right of Figure~\ref{fig:boxplots} summarize the optimizer statistics, showing little impact of the obstacle on the execution time.

\begin{table}[t]
	\centering
	\begin{tabular}{l | c c c}
		&\multicolumn{2}{c}{$\Delta t = 0.050\,$s} & $\Delta t = 0.150\,$s\\
		& Onboard & Offboard & Onboard\\
		\hline
		$k_\mathrm{max}$ & 1 & 1 & 1\\
		$l_\mathrm{max}$ & 2 & 30 & 6\\
		\# decision variables $n$ & 1504 & 1504 & 568 \\
		\hline
		$J$ & 8.102 & 4.737 & 14.08\\
		\# constraint violations$/t_n$ & 0.065 & 0.039 & 0.050 \\
		\# collisions & 3 & 0 & 1\\
		\hline
		Timely MPC steps & 99.50\,\% & 100\,\% & 100\,\%\\
		Async. $\boldsymbol{w}$ comm. steps & 0.69\,\% & 0\,\% & 6.19\,\%\\
		Async. $\bar{\boldsymbol{x}}$ comm. steps & 1.61\,\% & 0\,\% & 8.42\,\%\\
		Time spent computing & 39.32\,\%& 60.16\,\%& 13.37\,\%\\
		Time spent waiting/comm. & 60.68\,\% & 39.84\,\%& 86.66\,\%\\
	\end{tabular}
	\caption{Statistics for point-to-point transition without obstacle.}\label{tab:exp_results}
\end{table}

\subsection{Trajectory Tracking with Obstacle}
\begin{figure}
	\includegraphics[width = \columnwidth]{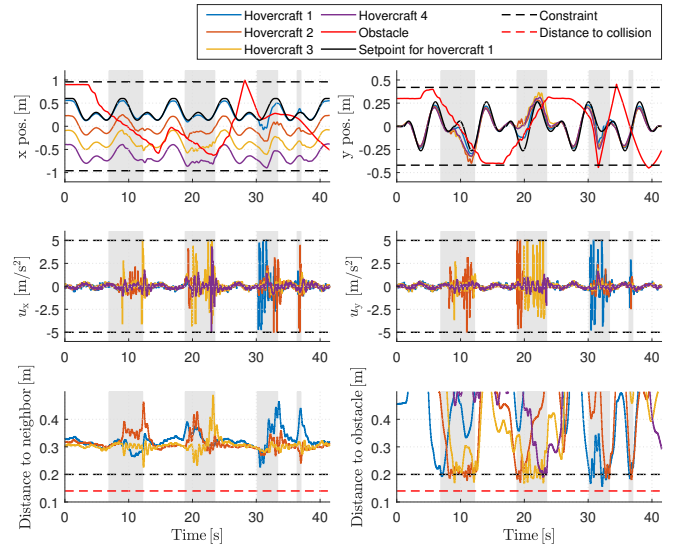}
	\caption{Trajectory tracking experiment with dynamic obstacle avoidance, offboard computation, and sampling interval $\Delta t = 50\,$ms. Areas shaded in gray mark stages where the obstacle inteferes with the tracking task. The DMPC controllers avoid collisions and continue tracking the trajectory once the obstacle has passed.}
	\label{fig:clover_obstacle}
\end{figure}

The third series of experiments tests the performance for more agile maneuvers.
The first hovercraft traverses a three-leaf clover path with parameter $\rho = 1$ in a lap time of $8\,$s, where $\rho$ is in the notation of~\cite{Dauer2013}. The remaining hovercraft maintain a constant distance to their neighbors and keep a line formation.
Meanwhile, an obstacle traverses the table and cuts through the formation several times such that the hovercraft have to interrupt tracking the precomputed trajectory in order to avoid collisions.
Figure~\ref{fig:clover_obstacle} shows the closed-loop position trajectories, control input, and distances to neighbors and the obstacle obtained with $k_\mathrm{max} = 1$ and $l_\mathrm{max} = 30$ iterations. 
The collision avoidance constraints between neighbors are always met, but there are some violations in the obstacle avoidance constraints.
Nevertheless, the hovercraft never actually collide with the obstacle due to the safety margin.
In general, the tracking performance is adequate, and the hovercraft follow the precomputed trajectory closely unless interrupted by the obstacle.

Finally, we note that the size of the air hockey table only permits the use of few hovercraft and the performance for larger swarms warrants further investigation. 
Recent numerical scalability studies show that the proposed dRTI scheme scales well at least up to $3.3\cdot10^5$ decision variables and several dozens of subsystems~\cite{Stomberg2024c}.

\section{Conclusion}

This paper has presented experimental results for cooperative DMPC based on dRTI.
The decentralized implementation was tested in formation control experiments with four hovercraft moving on an air hockey table.
The experimental hardware allowed to run the DMPC controllers offboard or onboard to test the performance on a variety of scenarios including point-to-point transitions, trajectory tracking, and dynamic obstacle avoidance.
For onboard computation, wireless communication was found to be the bottleneck preventing faster control sampling.
Future work will consider more efficient wireless communication to reduce onboard execution times as well as larger hovercraft swarms.

\clearpage

\renewcommand*{\bibfont}{\footnotesize}
\printbibliography

\end{document}